\documentclass[10pt, a4paper]{article}
\usepackage[final]{lrec2026} 

\usepackage[shortlabels]{enumitem}

\title{GS-BrainText: A Multi-Site Brain Imaging Report Dataset from Generation Scotland for Clinical Natural Language Processing Development and Validation}

\name{Beatrice Alex$^{1,2,3,9}$, Claire Grover$^{4}$, Arlene Casey$^{1,5}$, Richard Tobin$^{4}$, \\ {\bfseries \large Heather Whalley$^{6,7}$, William Whiteley$^{6,8}$}}

\address{$^{1}$Advanced Care Research Centre, University of Edinburgh, Edinburgh, UK \\
         $^{2}$School of Literatures, Languages and Cultures, University of Edinburgh, UK \\
         $^{3}$Edinburgh Futures Institute, University of Edinburgh, UK \\
         $^{4}$School of Informatics, University of Edinburgh, UK \\
         $^{5}$Usher Institute, University of Edinburgh, UK \\
         $^{6}$Institute for Neuroscience and Cardiovascular Research, University of Edinburgh, UK \\
         $^{7}$Generation Scotland, Institute of Genetics and Cancer, University of Edinburgh, UK \\
         $^{8}$NHS Lothian, Edinburgh, UK\\
         $^{9}$School of Mathematical and Computer Sciences, Heriot-Watt University, UK \\    
         Corresponding author: b.alex@ed.ac.uk \\}

\abstract{
We present GS-BrainText, a curated dataset of 8,511 brain radiology reports from the Generation Scotland cohort, of which 2,431 are annotated for 24 brain disease phenotypes. 
This multi-site dataset spans five Scottish NHS health boards and includes broad age representation (mean age 58, median age 53), making it uniquely valuable for developing and evaluating generalisable clinical natural language processing (NLP) algorithms and tools. Expert annotations were performed by a multidisciplinary clinical team using an annotation schema, with 10--100\% double annotation per NHS health board and rigorous quality assurance.  Benchmark evaluation using EdIE-R, an existing rule-based NLP system developed in conjunction with the annotation schema, revealed some performance variation across health boards (F1: 86.13--98.13), phenotypes (F1: 22.22--100) and age groups (F1: 87.01--98.13), highlighting critical challenges in generalisation of NLP tools. The GS-BrainText dataset addresses a significant gap in available UK clinical text resources and provides a valuable resource for the study of linguistic variation, diagnostic uncertainty expression and the impact of data characteristics on NLP system performance.\\ 
\newline
}

\begin{document}

\maketitleabstract

\section{Introduction}

Clinical natural language processing (NLP) can unlock valuable information from unstructured electronic health records, supporting research, clinical decision-making and healthcare quality improvement. However, the translation of NLP tools from research to clinical practice remains challenging, primarily due to limited access to diverse, representative datasets for development and validation.

Current clinical NLP research faces several critical limitations. First, most publicly available clinical text datasets originate from US healthcare systems and may not generalise to other countries, which differ in terminology, documentation practices and healthcare processes. Second, existing datasets often represent single institutions or narrow patient populations, limiting their utility for developing robust, generalisable NLP systems. Third, the highly sensitive nature of clinical data creates significant barriers to data sharing, hindering external validation of NLP tools.

We present GS-BrainText, a curated dataset of brain radiology reports from Generation Scotland, a broad population-based cohort spanning multiple Scottish NHS health boards. This dataset addresses key gaps in available clinical NLP resources by providing: (1) multi-site representation across various Scottish healthcare settings, (2) broad demographic coverage including age ranges typically under-represented in clinical datasets, (3) expert annotations for clinically important cerebrovascular phenotypes and (4) evidence of NLP system performance variation across phenotypes, sites and age ranges.

\section{Background}

\subsection{Generation Scotland}

Generation Scotland is a population-based health study of approximately 24,000 participants from the Scottish Family Health Study and the 21st Century Genetic Health study~\cite{smith2006generation,milbourn2024generation}. Established to investigate genetic and environmental contributions to common diseases, participants consented to data linkage and sharing of their health records, including radiology reports. The cohort spans multiple NHS health boards serving distinct geographic regions in Scotland with varying demographic characteristics. Unlike single-site or disease-specific cohorts, Generation Scotland represents a population-based sample that includes individuals (recruited between 2006 and 2010) across the full adult age spectrum.\footnote{Generation Scotland now also includes an additional 17,000 participants from across the whole of Scotland  (NextGenScot), including ones who are younger and from more health boards. Their data is not included in GS-BrainText.}

As part of routine clinical care, participants in Generation Scotland who underwent brain imaging (CT or MRI) have their radiology reports stored in NHS digital systems. These reports, authored by consultant radiologists, document findings observed in brain scans. Reports follow standard radiology reporting conventions, with sections for clinical history (provided by the referring clinician), observations and opinion. The length of the report varies from a few words to a maximum of 602 words (mean: 82 words) depending on the indication of the scan, the practice of the radiologist, and the complexity of the findings. Routine clinical reports are primarily written for clinical communication. The GS-BrainText dataset includes reports spanning 1994--2021, providing longitudinal imaging data for the Generation Scotland cohort.

A number of research study reports are also included in GS-BrainText but marked in the metadata to be excludable if necessary. Most of these study reports (784) are in the unannotated subset, with 14 in the annotated subset.

\subsection{Brain Disease Phenotypes}\label{subsection:phenotypes}

Phenotypes in clinical NLP refer to observable characteristics or traits that can be identified from medical records. In the context of brain radiology reports, phenotypes represent specific clinical findings or diagnostic categories that radiologists observe and document when interpreting brain imaging. For example, they include conditions such as ischaemic stroke (with specifications for location and time of occurrence), small vessel disease, atrophy and various types of tumours. 

Automated phenotype extraction from unstructured radiology text enables researchers to identify patient cohorts with specific characteristics, conduct large-scale epidemiological studies and investigate disease patterns that would be infeasible to detect through manual review alone. The accurate identification and classification of these phenotypes from free-text reports is essential for translating imaging findings into structured, searchable data that can support both clinical care and research.

We focus on 24 brain disease phenotypes with substantial clinical and research importance:

\begin{itemize}[label={}]
    \item \textbf{Stroke-Related Labels (11)}
        \begin{enumerate}[1.]
        \vspace{-0.2cm}
            \item Ischaemic stroke, deep, recent
            \vspace{-0.1cm}
            \item Ischaemic stroke, deep, old
            \vspace{-0.1cm}
            \item Ischaemic stroke, cortical, recent
            \vspace{-0.1cm}
            \item Ischaemic stroke, cortical, old
            \vspace{-0.1cm}
            \item Ischaemic stroke, underspecified
            \vspace{-0.1cm}
            \item Haemorrhagic stroke, deep, recent
            \vspace{-0.1cm}
            \item Haemorrhagic stroke, deep, old
            \vspace{-0.1cm}
            \item Haemorrhagic stroke, lobar, recent
            \vspace{-0.1cm}
            \item Haemorrhagic stroke, lobar, old
            \vspace{-0.1cm}
            \item Haemorrhagic stroke, underspecified
            \vspace{-0.1cm}
            \item Stroke, underspecified
        \end{enumerate}
    \item \textbf{Tumour-Related Labels (4)}
            \vspace{-0.1cm}
        \begin{enumerate}
            \addtocounter{enumi}{11}
            \item Tumour, meningioma
            \vspace{-0.1cm}
            \item Tumour, metastasis
            \vspace{-0.1cm}
            \item Tumour, glioma
            \vspace{-0.1cm}
            \item Tumour, other
        \end{enumerate}
    \item \textbf{Other Neurological Findings (9)}
    \vspace{-0.1cm}
        \begin{enumerate}
            \addtocounter{enumi}{15}
            \item Small vessel disease
            \vspace{-0.15cm}
            \item Atrophy
            \vspace{-0.15cm}
            \item Subdural haematoma
            \vspace{-0.15cm}
            \item Subarachnoid haemorrhage, aneurysmal
            \vspace{-0.5cm}
            \item Subarachnoid haemorrhage, other
            \vspace{-0.15cm}
            \item Microbleed, deep
            \vspace{-0.15cm}
            \item Microbleed, lobar
            \vspace{-0.15cm}
            \item Microbleed, underspecified
            \vspace{-0.15cm}
            \item Haemorrhagic transformation
        \end{enumerate}
\end{itemize}

These phenotypes are annotated as document-level labels (see Section~\ref{sub-section:annotation-schema}). They were first devised as part of earlier work on text mining brain imaging reports by \citet{alex2019text} using study specific data from one Scottish NHS health board (the Edinburgh Stroke Study, \cite{Jackson:2008}), which was motivated by the fact that existing medical ontological entries were not sufficiently fine-grained and phenotyping using image analysis was not sufficiently accurate yet. The document-level labels combine mark-up in the text itself for observations (like ischaemic stroke or tumour) as well as modifiers for location in the brain (deep/cortical/lobar) and time of occurrence (old/recent), creating a comprehensive taxonomy for stroke, tumour and other neurological findings from brain imaging reports. Automated extraction of these phenotypes enables large-scale research studies that would be infeasible through manual review alone \cite[e.g.,][]{camilleri2025large,iveson2025clinically}.

\subsection{Existing Datasets and Tools}

Despite the critical need for annotated clinical text datasets to develop and validate NLP systems, publicly available annotated brain imaging report datasets remain scarce. While some large-scale annotated radiology datasets exist, such as MIMIC-CXR~\cite{johnson2019mimic}, containing over 377,000 chest X-ray reports, and IU-XRay, the Indiana University Chest X-ray collection~\cite{demner2016preparing}, these focus on chest imaging rather than neuroimaging and originate from US healthcare systems. Linguistic conventions, terminology  and documentation practices in US radiology reports differ from UK NHS settings, limiting their utility for UK clinical deployment.

Recently, two major brain imaging datasets have been introduced. \citet{camilleri2025large} describe a large-scale national cohort from NHS Scotland (approx. 1.3M brain imaging reports/images linked to health systems data) accessible through the Brain Health Data Pilot with reports being labelled with EdIE-R system output.\footnote{\url{https://healthdatagateway.org/en/collection/158}} The Brain Imaging and Neurophysiology Database (BIND), approx.~85K radiology reports from over 1.8M neuroimaging scans across three US academic medical centres, used biomedical large language models (LLM) to extract structured clinical metadata from unstructured reports \cite{maschke2025The}. However, the authors explicitly note that due to LLM limitations and the absence of human-curated annotations, they "cannot guarantee complete accuracy of extracted information", positioning the metadata as a navigation tool rather than clinical ground truth. Both datasets provide scale but lack expert manual annotations of clinical phenotypes, limiting their utility for NLP systems requiring gold standard labels.

For expertly annotated brain imaging datasets, several smaller resources exist but with restricted access and limited scope. The Edinburgh Stroke Study dataset used to develop EdIE-R comprises 1,168 manually annotated reports~\cite{alex2019text}, and datasets used to develop ALARM+, ESPRESSO, and Sem-EHR remain internal to their respective institutions~\cite{schrempf2020paying,fu2019natural,rannikmae2021developing}. They provide high-quality expert annotations for specific phenotypes but are limited in scale, geographic diversity and availability to the broader research community.

This scarcity of shareable, expertly annotated, UK-based brain radiology datasets creates significant barriers to advancing clinical NLP research for neuroimaging applications. GS-BrainText addresses this gap by providing a multi-site, population-based resource with expert clinical gold annotations for UK NHS health boards.

\section{Data Collection and Curation}

\subsection{Report Acquisition}

We obtained access to brain imaging reports for Generation Scotland participants who underwent brain CT or MRI scans. Reports were retrieved through established data linkage protocols with NHS Scotland health boards: Fife, Lothian, Greater Glasgow and Clyde (GGC), Grampian and Tayside.

\subsection{Pre-processing Pipeline}

Raw reports and associated metadata were supplied in CSV format from each health board. We developed a standardised pre-processing pipeline comprising:

\noindent\textbf{Format conversion and structure identification:} Reports were converted from spreadsheets to XML format using components of the EdIE-R text processing pipeline~\cite{alex2019text}. During conversion, we automatically identified structural sections within reports, including clinical history and report body. Clinical history sections, which typically contain referring clinician notes rather than radiologist findings, were not annotated.

\noindent \textbf{Imaging modality labelling:} Reports were labelled by their modality (CT, MRI, etc.) in the XML.


Following this pre-processing, the final dataset, called the GS-BrainText data, comprises 8,511 scan reports (CT: 4362, MRI: 966, Other: 183) distributed across the five health boards: Tayside (3,747), GGC (3,309), Grampian (1,076), Lothian (269) and Fife (110). The distribution reflects the volume of reports originally provided by each board, with larger contributions from Tayside and GGC. From these available reports, we annotated 2,431 reports (CT: 1,487, MRI: 944), including all but 2 reports from Fife (108), approximately half from Lothian (143) and Grampian (521), a third from GGC (1,009) and a fifth from Tayside (650).

\section{Annotation Schema and Process}

\subsection{Annotation Schema Design}\label{sub-section:annotation-schema}

We employed a comprehensive annotation schema originally developed together with a neurologist and a radiologist for the development of the EdIE-R clinical NLP pipeline~\cite{alex2019text, wheater2019validated}. It comprises of text-level annotations of entities and modifiers, attributes of entities, relations as well as document-level labels:

\vspace{0.1cm}

\noindent \textbf{Text-level annotations} include (1) named entities representing clinical findings (e.g., ``ischaemic stroke'', ``small vessel disease'', ``atrophy''); (2) Location modifiers specifying anatomical regions (i.e., ``cortical'', ``lobar'' or ``deep'') and (3) temporal modifiers indicating timing (i.e., ``old'' or ``recent'').

\noindent \textbf{Attributes} are an additional annotation on entities reserved for negation (i.e., to signal that a type of stroke or tumour is not visible on the image.

\noindent \textbf{Relations} include links between modifiers and named entities (e.g., a link between ``old'' and ``ischaemic stroke'' to signal that this type of stroke was not recent or between ``deep'' and ``haemorrhagic stroke'' to signal the location of the haemorrhage. Entities can be related to multiple modifiers.

\noindent \textbf{Document-level labels} classify each report for all phenotypes (see Section~\ref{subsection:phenotypes}) in the context of all other annotations.

\vspace{0.1cm}

\begin{figure*}[!ht]
\begin{center}
\includegraphics[width=11.5cm]{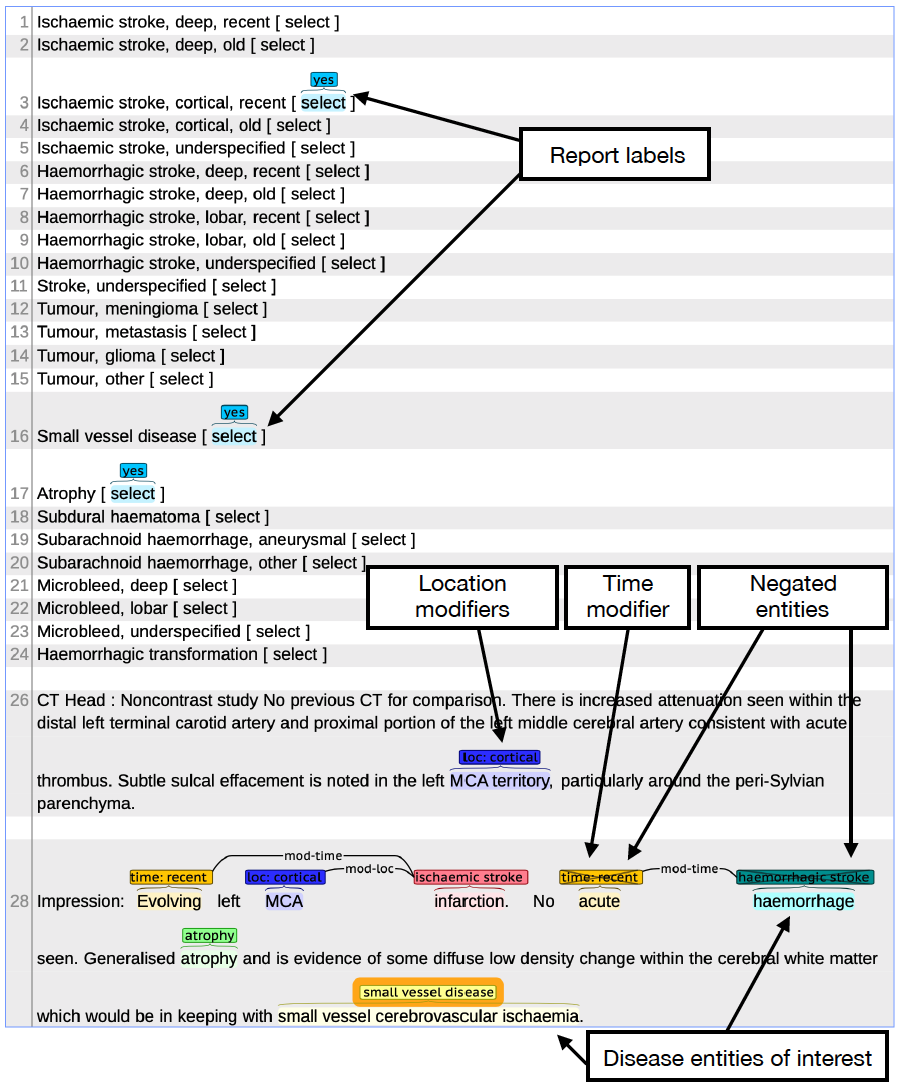}
\caption{A synthetic brain imaging report annotated with text- and document-level annotations.}
\label{fig:annotated-report}
\end{center}
\end{figure*}

The schema was designed to capture not only phenotype presence and absence but also critical contextual information that affects clinical interpretation. The annotation was carried out using the BRAT annotation tool \cite{stenetorp2012brat} and a synthetic annotated example is shown in Figure~\ref{fig:annotated-report}.

To ensure consistency and efficiency, annotation was performed as systematic correction of EdIE-R output rather than annotation from scratch. As a result, residual anchoring bias cannot be excluded; in particular, false negatives not surfaced by EdIE-R may be under-represented in the gold standard. Nevertheless, pre-annotation with EdIE-R substantially reduced annotation time, a pragmatic trade-off that is commonly adopted in clinical NLP data annotation work and one taken in light of EdIE-R's previously demonstrated performance on Scottish radiology report data (see Section~\ref{section:edie-r}.

\subsection{Annotation Team and Training}

Annotations were performed by a multidisciplinary team consisting of a consultant neurologist, a clinical fellow, a junior doctor, a clinical laboratory scientist and a medical student. All annotators underwent comprehensive training on 115 files from the Tayside and Fife subsets through an iterative process.

\subsection{Quality Assurance}

We implemented rigorous quality assurance to ensure annotation consistency:

\begin{itemize}
 \item Regular consensus meetings were held where annotators could discuss challenging cases, resolve queries and provide feedback to maintain consistency across all five annotators.
 \item Around 10\% of data was double-annotated across NHS GGC, Grampian and Tayside.
 \item 100\% of the NHS Lothian and Fife data was annotated by two or more annotators as we had less data available from these health boards.
 \item Inter-annotator agreement (IAA) was computed amongst the four less experienced annotators, i.e.\ not including the consultant neurologist (see Table~\ref{tab:iaa}).
 \item The final annotated dataset comprises single-annotated files and, for double-annotated files, annotations selected based on annotator expertise and consistency.
\end{itemize}

Table~\ref{tab:iaa} presents IAA statistics for document-level labels (phenotypes). F1-scores range from 83.74 to 95.65 across phenotypes and health boards, indicating reasonably high or very high agreement.

\begin{table*}
\begin{center}
\begin{tabular}{|l|c|c|c|c|c|}
\hline
\textbf{Annotator pairs}       & \textbf{Fife}  & \textbf{GGC} & \textbf{Grampian} & \textbf{Lothian} & \textbf{Tayside}\\
\hline 
 A1-A2 & 88.15 (108) &   -    &   -     &   -      & -\\
 A1-A3 & 87.50 (108) &   -    &   -     &   -      & -\\
 A1-A4 & 89.90 (108) &   -    &   -     &   -      & -\\
 A2-A3 & 83.58 (108) & 91.18 (53) &   -     & 86.51 (143)  & 90.91 (35) \\
 A2-A4 & 90.82 (108) &   -    & 95.65 (26) &   -      & -\\
 A3-A4 & 86.17 (108) & 86.49 (53) & 91.67 (24) &   -      & 87.50 (30)\\
 \hline
\end{tabular}
\caption{Pairwise inter-annotator agreement (IAA) reported in F1 (and number of reports in parentheses) for phenotype annotations in GS-BrainText across five Scottish NHS health boards (GGC = Greater Glasgow and Clyde). All annotators annotated all the reports in Fife and two annotated all the reports in Lothian. IAA for other boards was computed on smaller subsets with two pairwise comparisons each.}
\label{tab:iaa}
\end{center}
\end{table*}

 IAA was assessed at both the text level (named entities, negation attributes, modifiers and relations) and the document level (phenotype labels) in GS-BrainText. However, given this paper's focus on dataset utility for phenotype extraction, we only report document-level agreement statistics. Comprehensive text-level annotation statistics, albeit for other datasets, can be found in~\citet{alex2019text} and \citet{wheater2019validated}.

\section{Annotated Dataset Characteristics}

\subsection{Size and Composition}

The annotated GS-BrainText comprises 2,431 radiology reports across five NHS health boards. Table~\ref{tab:distribution} shows the distribution of reports and phenotype prevalence by health board and demonstrates substantial variation in frequencies.\footnote{We do not provide a fixed data split, as the annotated subset was used solely for evaluation and several phenotypes are rare in the data. We therefore recommend future work adopt k-fold cross-validation to determine performance of the entire dataset.} Small vessel disease (n=545) and atrophy (n=441) were the most common findings, together accounting for over 53\% of all labels. Ischaemic stroke phenotypes show reasonable representation (n=381 total) and were  more common than haemorrhagic stroke (n=84 total), consistent with known stroke epidemiology. Several clinically important phenotypes were rare in the dataset: microbleeds (n=12), haemorrhagic transformation (n=10) and gliomas (n=8). This imbalanced distribution reflects both the population-based nature of the cohort and the relative epidemiological frequencies of these conditions, but poses challenges for evaluating NLP system performance on rare phenotypes.

The distribution varies also substantially by health board, with GGC contributing the largest absolute numbers across most phenotypes due to its larger sample size. Rarer phenotypes like microbleeds appear almost exclusively in data from NHS GGC and Grampian.

\begin{table*}[t]
\begin{center}
\begin{tabular}{|l|r|r|r|r|r|r|}
\hline
\textbf{Phenotype} & \textbf{Fife} & \textbf{GGC} & \textbf{Grampian} & \textbf{Lothian} & \textbf{Tayside} & \textbf{Total}\\
\hline
Ischaemic\_stroke\_deep\_recent & 0 & 9 & 9 & 3 & 0 & 21\\
Ischaemic\_stroke\_deep\_old & 6 & 78 & 55 & 15 & 25 & 179\\
Ischaemic\_stroke\_cortical\_recent & 6 & 22 & 9 & 5 & 5 & 47\\
Ischaemic\_stroke\_cortical\_old & 6 & 45 & 24 & 5 & 8 & 88\\
Ischaemic\_stroke\_underspecified & 4 & 17 & 11 & 0 & 14 & 46\\
Haemorrhagic\_stroke\_deep\_recent & 0 & 0 & 1 & 1 & 0 & 2\\
Haemorrhagic\_stroke\_deep\_old & 1 & 3 & 1 & 0 & 2 & 7\\
Haemorrhagic\_stroke\_lobar\_recent & 1 & 4 & 1 & 1 & 3 & 10\\
Haemorrhagic\_stroke\_lobar\_old & 1 & 8 & 1 & 0 & 1 & 11\\
Haemorrhagic\_stroke\_underspecified & 6 & 28 & 11 & 3 & 6 & 54\\
Stroke\_underspecified & 2 & 3 & 2 & 0 & 1 & 8\\
Tumour\_meningioma & 7 & 18 & 18 & 2 & 3 & 48\\
Tumour\_metastasis & 0 & 22 & 6 & 4 & 2 & 34\\
Tumour\_glioma & 0 & 4 & 3 & 1 & 0 & 8\\
Tumour\_other & 3 & 24 & 10 & 5 & 9 & 51\\
Small\_vessel\_disease & 26 & 250 & 178 & 42 & 49 & 545\\
Atrophy & 32 & 215 & 114 & 21 & 59 & 441\\
Subdural\_haematoma & 0 & 36 & 25 & 7 & 0 & 68\\
Subarachnoid\_haemorrhage\_aneurysmal & 0 & 2 & 4 & 1 & 0 & 7\\
Subarachnoid\_haemorrhage\_other & 1 & 9 & 15 & 5 & 3 & 33\\
Microbleed\_deep & 0 & 1 & 1 & 0 & 0 & 2\\
Microbleed\_lobar & 0 & 3 & 4 & 0 & 0 & 7\\
Microbleed\_underspecified & 0 & 3 & 0 & 0 & 0 & 3\\
Haemorrhagic\_transformation & 5 & 4 & 1 & 0 & 0 & 10\\
\hline
\end{tabular}
\caption{Distribution of phenotype prevalence across five Scottish NHS health boards in the annotated subset of GS-BrainText (GGC = Greater Glasgow and Clyde).}
\label{tab:distribution}
\end{center}
\end{table*}


\subsection{Demographic Distribution}

The annotated dataset contains 2,431 scan reports, of which 1,433 are of female and 988 of male patients, matching the overall Generation Scotland cohort gender distribution: 59\% female and 41\% male \cite{smith2012cohort}.

Of the annotated reports, 644 are from patients under 50, 1,063 from those aged 50–69, and 704 from those aged over 70.

\subsection{Linguistic Characteristics}

\citet{casey2023understanding} previously identified important linguistic characteristics in a subset of the GS-BrainText dataset related to diagnostic uncertainty in radiological reporting. Their analysis showed that scan reports of younger patients frequently employ hedging language, particularly for cerebrovascular findings that are less expected in this age group. Phrases such as ``this may possibly be an ischaemic infarct'' appear more frequently in reports of patients under 50 years compared to those over 70. Uncertainty expressions pose challenges both for manual annotation and NLP systems.

\section{Benchmark Evaluation}
In prior work, \citet{casey2023understanding} compared the performance of four NLP systems on CT scans in GS-BrainText: EdIE-R~\cite{alex2019text, wheater2019validated}, ALARM+~\cite{schrempf2020paying, schrempf2021templated}, ESPRESSO~\cite{fu2019natural, fu2020assessment} and Sem-EHR~\cite{rannikmae2021developing}. These systems represent diverse approaches (rule-based and neural) and training or development data sources.  Evaluation was restricted to a subset of three overlapping phenotypes (ischaemic stroke, small vessel disease and atrophy).

In the following section, we will lay out the performance for EdIE-R across all 24 phenotypes on the full manually annotated GS-BrainText gold data (CT and MRI).  The aim is to demonstrate the usefulness of the GS-BrainText dataset providing a baseline for the entire phenotype annotation scheme for future research using machine learning or large language model approaches.

\subsection{System Description}\label{section:edie-r}

EdIE-R (short for Edinburgh Information Extraction for Radiology) is a rule-based system developed on the Edinburgh Stroke Study (ESS) data from NHS Lothian \cite{alex2019text} and subsequently refined and evaluated on NHS Tayside data \cite{wheater2019validated}.\footnote{EdIE-R is available at \url{https://www.ltg.ed.ac.uk/software/edie-r}.} After some initial preprocessing and linguistic analysis, it performs named entity recognition (NER), negation detection and relation extraction before conducting document-level classification. Its NER and negation detection components were robustly evaluated on ESS and Tayside data \cite{grivas2020not,sykes2021comparison}.

\subsection{Evaluation}

Table~\ref{tab:performance} presents EdIE-R's precision, recall and F1 scores across all 24 phenotypes on the manually annotated GS-BrainText data.\footnote{In clinical research, precision is also called positive predictive value and recall is known as sensitivity.} EdIE-R demonstrates high overall performance on this dataset, achieving a micro-average F1 score of 88.82. However, micro-averaging is dominated by high-frequency phenotypes; we therefore also report macro-averaged precision, recall and F1 of 74.22, 84.78 and 77.47 respectively, which weights all phenotypes equally and reflects performance across the full range of phenotype frequencies.

System performance varies considerably by phenotype type and frequency, achieving best results on high-frequency phenotypes: small vessel disease (F1=95.6), atrophy (F1=95.69) and subdural haematoma (F1=91.55). For ischaemic stroke subtypes, performance ranges from 74.42 (deep, recent) to 91.3 (cortical, recent), with the underspecified category performing notably lower (F1=60) due to reduced precision (51.56). Microbleeds (deep and lobar) achieve perfect scores (F1=100), though their extreme rarity (n=2 and n=7) limits the reliability of these estimates. The poorest performance is obtained for subarachnoid haemorrhage, aneurysmal (F1=22.22), reflecting both the low frequency of this phenotype (n=7) and its very low recall (14.29).

Individual results for low-support phenotypes (n<30 in the Total column of Table~\ref{tab:distribution}) should be interpreted with caution, as scores may not be reliable estimates of generalised system performance.

\begin{table*}[t]
\centering
\small
\begin{tabular}{|l|rrr|}
\hline
\textbf{Phenotype} & \textbf{Precision} & \textbf{Recall} & \textbf{F1} \\
\hline
Ischaemic\_stroke\_deep\_recent & 72.73 & 76.19 & 74.42\\
Ischaemic\_stroke\_deep\_old & 84.58 & 94.97 & 89.47\\
Ischaemic\_stroke\_cortical\_recent & 93.33 & 89.36 & 91.30\\
Ischaemic\_stroke\_cortical\_old & 86.02 & 90.91 & 88.40\\
Ischaemic\_stroke\_underspecified & 51.56 & 71.74 & 60.00\\
Haemorrhagic\_stroke\_deep\_recent & 66.67 & 100.00 & 80.00\\
Haemorrhagic\_stroke\_deep\_old & 75.00 & 42.86 & 54.55\\
Haemorrhagic\_stroke\_lobar\_recent & 77.78 & 70.00 & 73.68\\
Haemorrhagic\_stroke\_lobar\_old & 72.73 & 72.73 & 72.73\\
Haemorrhagic\_stroke\_underspecified & 57.14 & 96.30 & 71.72\\
Stroke\_underspecified & 36.84 & 87.50 & 51.85\\
Tumour\_meningioma & 82.76 & 100.00 & 90.57\\
Tumour\_metastasis & 64.00 & 94.12 & 76.19\\
Tumour\_glioma & 66.67 & 75.00 & 70.59\\
Tumour\_other & 54.05 & 78.43 & 64.00\\
Small\_vessel\_disease & 93.66 & 97.61 & 95.60\\
Atrophy & 98.33 & 93.20 & 95.69\\
Subdural\_haematoma & 87.84 & 95.59 & 91.55\\
Subarachnoid\_haemorrhage\_aneurysmal & 50.00 & 14.29 & 22.22\\
Subarachnoid\_haemorrhage\_other & 63.27 & 93.94 & 75.61\\
Microbleed\_deep & 100.00 & 100.00 & 100.00\\
Microbleed\_lobar & 100.00 & 100.00 & 100.00\\
Microbleed\_underspecified & 75.00 & 100.00 & 85.71\\
Haemorrhagic\_transformation & 71.43 & 100.00 & 83.33\\
\hline
Micro-average & 85.03 & 92.95 & 88.82\\
\hline
Macro-average & 74.22 & 84.78 & 77.47\\
\hline
\end{tabular}
\caption{Precision, recall and F1 scores for the EdIE-R NLP system on GS-BrainText for individual phenotypes as well as micro- and macro-averaged performance scores.}
\label{tab:performance}
\end{table*}


\subsection{Performance by Health Board}

\begin{table}[t]
\centering
\small
\begin{tabular}{|l|c|c|c|}
\hline
\textbf{Health Board} & \textbf{Precision} & \textbf{Recall} & \textbf{F1}\\
\hline
Fife & 98.13 & 98.13 & 98.13 \\
GGC & 82.18 & 92.45 & 87.01 \\
Grampian & 87.12 & 91.27 & 89.15 \\
Lothian & 94.44 & 98.35 & 96.36 \\
Tayside & 80.09 & 93.16 & 86.13 \\
\hline
\end{tabular}
\caption{EdIE-R performance in precision, recall and micro-average F1-scores for GS-BrainText per NHS health board.}
\label{tab:board_performance}
\end{table}

EdIE-R performance across health boards (see Table~\ref{tab:board_performance}) reveals some variation, with F1 scores ranging from 86.13 (Tayside) to 98.13 (Fife). The system achieves highest performance on Lothian (96.36) and Fife (98.13). Notably, Lothian data (though unlikely the specific reports in GS-BrainText) were used in EdIE-R's original development on the Edinburgh Stroke Study~\cite{alex2019text}, while data from Fife was not used when devising its rules. Performance is lower on other health boards: GGC (87.01), Grampian (89.15) and Tayside (86.13). The reduced performance on Tayside is particularly noteworthy, as similar NHS Tayside data was used for system refinement~\cite{wheater2019validated}. The variation in F1 scores demonstrates the impact of local reporting conventions and data heterogeneity on rule-based system performance, even within a single national healthcare system.

\subsection{Performance by Age Group}

\begin{table}[t]
\centering
\small
\begin{tabular}{|l|c|c|c|}
\hline
\textbf{Age Group} & \textbf{Precision} & \textbf{Recall} & \textbf{F1} \\
\hline
<50 & 69.78 & 87.39 & 77.60\\
50-69 & 81.24 & 92.76 & 86.62\\
70+ & 89.78 & 93.70 & 91.70\\
\hline
\end{tabular}
\caption{EdIE-R performance in precision, recall and micro-average F1-scores by age group.}
\label{tab:age_performance}
\end{table}

Performance increased with age (ranging between F1=77.60 for <50 and F1=91.7 for 70+), corresponding to phenotype coverage of 16/24 for people under 50, 24/24 for people aged 50-69, and 23/24 for those aged over 70 (see Table~\ref{tab:age_performance}).

\section{Discussion and Conclusion}

We present BS-BrainText, a dataset of 8,511 brain radiology reports from Generation Scotland, of which 2,431 are annotated for 24 cerebrovascular phenotypes across five Scottish NHS health boards. Benchmark evaluation of the EdIE-R NLP system reveals strong overall performance (F1=88.82) with substantial variation across phenotypes (22.22--100), health boards (86.13--98.13) and age (77.6--91.7), highlighting challenges for clinical NLP deployment. Performance exceeds 95 in F1 for common phenotypes, but drops for rare ones and differs across NHS sites, even within a single national healthcare system. 

The observed variation highlights critical factors affecting clinical NLP robustness. Local reporting conventions substantially impact performance: EdIE-R achieves 96.36 on Lothian data similar to what was used in its development, yet performs at 86--89 on data from some other boards. Interestingly, performance on Fife (98.13) exceeds that on Tayside (86.13) despite similar Tayside data having been used for system refinement. Its performance variation of up to 12 points across sites underscores the challenges of developing truly generalisable clinical NLP systems. 


These findings carry important implications for clinical deployment. "Out-of-the-box" application of NLP systems to new healthcare settings, even within the same country, carries some risk of performance degradation. Organisations adopting NLP tools should therefore conduct local validation before deployment and monitor performance across demographic and clinical subgroups. For researchers, our results underscore the value of multi-site datasets for developing robust systems, as training on single-institution data does not prepare systems for linguistic and clinical variation encountered across sites.

GS-BrainText, containing  expert annotations across 24 clinically important phenotypes, provides a valuable resource for advancing robust and generalisable clinical NLP, enabling researchers to develop systems that can handle the linguistic variation, demographic diversity and multi-site heterogeneity encountered in real-world healthcare settings. Through controlled access, this dataset can support the development of next-generation clinical NLP tools that maintain reliable performance across the diverse contexts of clinical practice. 

With EdIE-R, we contribute a strong baseline (micro-average F1=88.82), demonstrating that well-engineered rule-based approaches can achieve competitive performance on clinical phenotyping tasks while providing interpretable predictions for benchmarking new approaches.

\section{Limitations and Future Work}

While GS-BrainText addresses important gaps in available clinical NLP resources, several limitations and opportunities for future work should be noted:
\vspace{0.1cm}

\noindent \textbf{Geographic scope:} The dataset is limited to Scottish NHS health boards and may not represent clinical reporting practices in other UK regions or international healthcare systems. Variations in terminology, reporting conventions and clinical workflows may limit generalisability beyond Scotland, though the multi-site nature within Scotland provides some indication of cross-institutional robustness.

\noindent \textbf{Imaging modality:} The dataset includes both CT (1,487 annotated, 4,362 total) and MRI (944 annotated, 3,966 total) reports. The predominance of CT in the annotated subset reflects its prevalence in acute clinical settings, though the substantial MRI representation (39\% of annotated reports) enables investigation of cross-modality language variation. Terminology and reporting conventions may differ between CT and MRI, presenting both challenges and opportunities for developing robust cross-modality systems.

\noindent \textbf{Phenotype coverage:} Although we annotated 24 clinically important phenotypes, many other neurological findings documented in radiology reports are not captured by our annotation schema. The schema focuses on cerebrovascular pathology and common structural abnormalities, excluding conditions such as infectious processes and traumatic injuries beyond subdural haematoma.

\noindent \textbf{Class imbalance:} Substantial imbalance exists in phenotype frequencies, with some conditions (small vessel disease, atrophy) well-represented while others (microbleeds, certain tumour subtypes) being extremely rare. This imbalance, while reflecting real-world epidemiology, limits the ability to robustly evaluate NLP system performance on rare but clinically important phenotypes.

\noindent \textbf{Annotation reporting:} Document-level labels represent the primary annotation layer reported on in this paper, with text-level annotations (named entities, relations, temporal and location modifiers and negation) not systematically reported here. While phenotype annotation aligns with typical use cases for cohort identification and epidemiological research, researchers interested in developing entity recognition or relation extraction components can acquire additional detailed annotation statistics from the GS-BrainText data itself.

\noindent \textbf{Temporal factors:} Reports span over a large time frame (1994--2021; most common scan date in 2016) during which reporting practices, imaging technology and clinical guidelines have evolved. We do not systematically analyse temporal trends in language use or diagnostic patterns, though such changes may contribute to observed performance variation across subsets of the data and can be explored in future research.

\noindent \textbf{Population characteristics:} Generation Scotland is a volunteer cohort that may not fully represent the general Scottish population. Participants willing to engage in long-term clinical research may differ from non-participants in their readiness to seek medical help, their socioeconomic status or their baseline health. The population-based sampling reduces some selection biases present in disease-specific cohorts, but perfect representativeness cannot be assumed.

\noindent \textbf{NLP system coverage:} We evaluated only EdIE-R, a rule-based system. Modern contextual models may improve on some challenges, but evidence suggests site-specific variation persists across NLP architectures (Casey et al., 2023), indicating that architectural advances alone cannot eliminate the need for local validation.

Despite these limitations, GS-BrainText represents a valuable available resources for clinical NLP research on brain imaging reports, providing multi-site, expertly annotated data with comprehensive baseline performance reporting and interesting opportunities for future research.

\section{Data Access} 
GS-BrainText is available via Generation Scotland's controlled access process: \url{https://www.ed.ac.uk/generation-scotland/for-researchers/access}.

\section{Ethics Statement}

This research was conducted under ethical approvals for the Generation Scotland cohort. GS has ethical approval for the SFHS study (05/S1401/89) and 21CGH study (06/S1401/27) and both studies are now part of a Research Tissue Bank (20-ES-0021). Ethical approval for the GS:SFHS study was obtained from the Tayside Committee on Medical Research Ethics (on behalf of the National Health Service). Ethical approval for the GS:21CGH study was obtained from the Scotland A Research Ethics Committee. Patients/participants provided their written informed consent to participate in this study.

\section{Acknowledgements}

We thank the Generation Scotland participants and the Generation Scotland team. Generation Scotland received core support from the Chief Scientist Office of the Scottish Government Health Directorates [CZD/16/6] and the Scottish Funding Council [HR03006]. Genotyping of the GS:SFHS samples was carried out by the Genetics Core Laboratory at the Wellcome Trust Clinical Research Facility, Edinburgh, Scotland, was funded by the Medical Research Council UK and has been and is currently funded by the Wellcome Trust (STratifying Resilience and Depression Longitudinally  (STRADL) Reference 104036/Z/14/Z and Generation Scotland: NextGenScot Reference 216767/Z/19/Z). 

We also thank the annotators (Liam Lee, Michael Walsh, Freya Pellie, Karen Ferguson and William Whiteley) who contributed to creating the GS-BrainText gold standard labels.

Authors (initials in brackets) were supported by:
\begin{itemize}
    \item Turing Fellowships (B.A. and C.G.) and a Turing project (B.A. and A.C.) funded by the Alan Turing Institute (EPSRC grant EP/N510129/1) 
    \item Legal\& General Group as part of the Advanced Care Research Centre (B.A. and A.C.) 
    \item The AIM-CISC project funded by the National Institute for Health Research (NIHR202639; B.A.) 
    \item An MRC Clinician Scientist Award (G0902303; W.W.) 
    \item A Scottish Senior Clinical Fellowship (CAF/17/01; W.W.)
The funders had no role in the conduct of the study, interpretation or the decision to submit for publication. The views expressed are those of the authors and not necessarily those of the funders.
\end{itemize}
 

\section{Bibliographical References}\label{sec:reference}

\bibliographystyle{lrec2026-natbib}
\bibliography{bibliography}

@article{smith2006generation,
  title={Generation Scotland: the Scottish Family Health Study; a new resource for researching genes and heritability},
  author={Smith, Blair H and Campbell, Harry and Blackwood, Douglas and Connell, John and Connor, Mike and Deary, Ian J and Dominiczak, Anna F and Fitzpatrick, Bridie and Ford, Ian and Jackson, Cathy and others},
  journal={BMC medical genetics},
  volume={7},
  number={1},
  pages={74},
  year={2006},
  publisher={Springer},
  url={https://doi.org/10.1186/1471-2350-7-74}
}

@article{smith2012cohort,
    author = {Smith, Blair H and Campbell, Archie and Linksted, Pamela and Fitzpatrick, Bridie and Jackson, Cathy and Kerr, Shona M and Deary, Ian J and MacIntyre, Donald J and Campbell, Harry and McGilchrist, Mark and Hocking, Lynne J and Wisely, Lucy and Ford, Ian and Lindsay, Robert S and Morton, Robin and Palmer, Colin N A and Dominiczak, Anna F and Porteous, David J and Morris, Andrew D},
    title = {Cohort Profile: Generation Scotland: Scottish Family Health Study (GS:SFHS). The study, its participants and their potential for genetic research on health and illness},
    journal = {International Journal of Epidemiology},
    volume = {42},
    number = {3},
    pages = {689-700},
    year = {2012},
    month = {07},
    issn = {0300-5771},
    doi = {10.1093/ije/dys084},
    url = {https://doi.org/10.1093/ije/dys084},
    eprint = {https://academic.oup.com/ije/article-pdf/42/3/689/18481950/dys084.pdf},
}

@article{alex2019text,
  title={Text mining brain imaging reports},
  author={Alex, Beatrice and Grover, Claire and Tobin, Richard and Sudlow, Cathie and Mair, Grant and Whiteley, William},
  journal={Journal of Biomedical Semantics},
  volume={10},
  number={Supplement 1},
  pages={23},
  year={2019},
  url = {https://doi.org/10.1186/s13326-019-0211-7},
  publisher={Springer}
}

@article{camilleri2025large,
  title={{A large dataset of brain imaging linked to health systems data: the curation and access to a whole system national cohort from NHS Scotland}},
  author={Camilleri, Michael PJ and Gouzou, Dorian and Al-Wasity, Salim and Mookiah, Muthu RK and Hernandez, Mar{\'\i}a Valdes and Alex, Bea and Tsaftaris, Sotirios A and Brooks, Andrew and MacLeod, Ruairidh and Wu, Honghan and others},
  journal={medRxiv},
  pages={2025--10},
  year={2025},
  publisher={Cold Spring Harbor Laboratory Press},
  url={https://doi.org/10.1101/2025.10.21.25338469}
}

@article{wheater2019validated,
  title={{A validated natural language processing algorithm for brain imaging phenotypes from radiology reports in UK electronic health records}},
  author={Wheater, Emily and Mair, Grant and Sudlow, Cathie and Alex, Beatrice and Grover, Claire and Whiteley, William},
  journal={BMC Medical Informatics and Decision Making},
  volume={19},
  number={1},
  pages={184},
  year={2019},
  publisher={Springer},
  url={https://doi.org/10.1186/s12911-019-0908-7}
}

@inproceedings{stenetorp2012brat,
  title={{BRAT: a web-based tool for NLP-assisted text annotation}},
  author={Stenetorp, Pontus and Pyysalo, Sampo and Topi{\'c}, Goran and Ohta, Tomoko and Ananiadou, Sophia and Tsujii, Jun’ichi},
  booktitle={Proceedings of the Demonstrations at the 13th Conference of the European Chapter of the Association for Computational Linguistics},
  pages={102--107},
  year={2012},
  url={https://aclanthology.org/E12-2021/}
}

@article{casey2023understanding,   
AUTHOR={Casey, Arlene  and Davidson, Emma  and Grover, Claire  and Tobin, Richard  and Grivas, Andreas  and Zhang, Huayu  and Schrempf, Patrick  and O’Neil, Alison Q.  and Lee, Liam  and Walsh, Michael  and Pellie, Freya  and Ferguson, Karen  and Cvoro, Vera  and Wu, Honghan  and Whalley, Heather  and Mair, Grant  and Whiteley, William  and Alex, Beatrice }, 
TITLE={{Understanding the performance and reliability of NLP tools: a comparison of four NLP tools predicting stroke phenotypes in radiology reports}},     
JOURNAL={Frontiers in Digital Health},
VOLUME={5},
YEAR={2023},
URL={https://www.frontiersin.org/journals/digital-health/articles/10.3389/fdgth.2023.1184919},
DOI={10.3389/fdgth.2023.1184919},
ISSN={2673-253X}}

@inproceedings{schrempf2020paying,
  title={Paying per-label attention for multi-label extraction from radiology reports},
  author={Schrempf, Patrick and Watson, Hannah and Mikhael, Shadia and Pajak, Maciej and Falis, Mat{\'u}{\v{s}} and Lisowska, Aneta and Muir, Keith W and Harris-Birtill, David and O’Neil, Alison Q},
  booktitle={International Workshop on Interpretability of Machine Intelligence in Medical Image Computing},
  pages={277--289},
  year={2020},
  organization={Springer},
  url={https://doi.org/10.1007/978-3-030-61166-8_29}
}

@article{schrempf2021templated,
  title={Templated text synthesis for expert-guided multi-label extraction from radiology reports},
  author={Schrempf, Patrick and Watson, Hannah and Park, Eunsoo and Pajak, Maciej and MacKinnon, Hamish and Muir, Keith W and Harris-Birtill, David and O’Neil, Alison Q},
  journal={Machine Learning and Knowledge Extraction},
  volume={3},
  number={2},
  pages={299--317},
  year={2021},
  publisher={MDPI},
  url={https://www.mdpi.com/1046980}
}

@article{fu2019natural,
  title={Natural language processing for the identification of silent brain infarcts from neuroimaging reports},
  author={Fu, Sunyang and Leung, Liwei Y and Wang, Yiqing and Raulli, Anne-Olivia and Kallmes, David F and Kinsman, Kerry A and Nelson, Kristin B and Clark, Thomas K and Luetmer, Patrick H and Kingsbury, Peter R and others},
  journal={JMIR Medical Informatics},
  volume={7},
  number={2},
  pages={e12109},
  year={2019},
  publisher={JMIR Publications Inc., Toronto, Canada},
  doi={10.2196/12109}
}

@article{fu2020assessment,
  title={Assessment of the impact of {EHR} heterogeneity for clinical research through a case study of silent brain infarction},
  author={Fu, Sunyang and Leung, Liwei Y and Raulli, Anne-Olivia and Kallmes, David F and Kinsman, Kerry A and Nelson, Kristin B and Clark, Thomas K and Luetmer, Patrick H and Kingsbury, Peter R and Kent, David M and others},
  journal={BMC Medical Informatics and Decision Making},
  volume={20},
  number={1},
  pages={60},
  year={2020},
  publisher={Springer},
  doi={10.1186/s12911-020-1072-9}
}

@article{rannikmae2021developing,
  title={Developing automated methods for disease subtyping in {UK Biobank}: An exemplar study on stroke},
  author={Rannikmäe, Kristiina and Wu, Honghan and Tominey, Sophie and Whiteley, William and Allen, Naomi and Sudlow, Cathie {UK Biobank Stroke Study}},
  journal={BMC Medical Informatics and Decision Making},
  volume={21},
  number={191},
  year={2021},
  publisher={Springer},
  doi={10.1186/s12911-021-01556-0}
}

@inproceedings{grivas2020not,
  title={Not a cute stroke: analysis of rule-and neural network-based information extraction systems for brain radiology reports},
  author={Grivas, Andreas and Alex, Beatrice and Grover, Claire and Tobin, Richard and Whiteley, William},
  booktitle={The 11th International Workshop on Health Text Mining and Information Analysis at EMNLP 2020},
  pages={24--37},
  year={2020},
  organization={Association for Computational Linguistics},
  url={https://aclanthology.org/2020.louhi-1.4/}
}

@article{sykes2021comparison,
  title={Comparison of rule-based and neural network models for negation detection in radiology reports},
  author={Sykes, Dominic and Grivas, Andreas and Grover, Claire and Tobin, Richard and Sudlow, Cathie and Whiteley, William and McIntosh, Andrew and Whalley, Heather and Alex, Beatrice},
  journal={Natural Language Engineering},
  volume={27},
  number={2},
  pages={203--224},
  year={2021},
  publisher={Cambridge University Press},
  url={https://dx.doi.org/10.1017/s1351324920000509}
}

@article{Jackson:2008,
	author  = {Jackson, Caroline and Crossland, Laura and Dennis, Michele and Wardlaw, Joanna and Sudlow, Cathie}, 
	title   = {Assessing the impact of the requirement for explicit consent in a hospital-based stroke study},
	journal = {QJM: Monthly Journal of the Association of Physicians}, 
	year  = {2008},
	  month = {}, 
	  volume= {101}, 
	  number= {4}, 
	  pages = {281-289},
      url={https://pubmed.ncbi.nlm.nih.gov/18281363/}
}

@article{iveson2025clinically,
	author = {Iveson, Matthew H and Mukerjee, Mome and Davidson, Emma M and Zhang, Huayu and Sherlock, Laura and Ball, Emily L and Mair, Grant and Hosking, Alice and Whalley, Heather and Poon, Michael T C and Wardlaw, Joanna M and Kent, David and Tobin, Richard and Grover, Claire and Alex, Beatrice and Whiteley, William N},
	title = {Clinically reported covert cerebrovascular disease and risk of neurological disease: a whole-population cohort of 395,273 people using natural language processing},
	elocation-id = {2025.06.12.25329472},
	year = {2025},
	doi = {10.1101/2025.06.12.25329472},
	publisher = {Cold Spring Harbor Laboratory Press},
	URL = {https://www.medrxiv.org/content/early/2025/06/23/2025.06.12.25329472},
	eprint = {https://www.medrxiv.org/content/early/2025/06/23/2025.06.12.25329472.full.pdf},
	journal = {medRxiv}
}

@article{johnson2019mimic,
  title={{MIMIC-CXR}, a de-identified publicly available database of chest radiographs with free-text reports},
  author={Johnson, Alistair EW and Pollard, Tom J and Berkowitz, Seth J and Greenbaum, Nathaniel R and Lungren, Matthew P and Deng, Chih-ying and Mark, Roger G and Horng, Steven},
  journal={Scientific Data},
  volume={6},
  number={1},
  pages={317},
  year={2019},
  publisher={Nature Publishing Group},
  doi={10.1038/s41597-019-0322-0}
}

@article{demner2016preparing,
  title={Preparing a collection of radiology examinations for distribution and retrieval},
  author={Demner-Fushman, Dina and Kohli, Marc D and Rosenman, Marc B and Shooshan, Sonya E and Rodriguez, Laritza and Antani, Sameer and Thoma, George R and McDonald, Clement J},
  journal={Journal of the American Medical Informatics Association},
  volume={23},
  number={2},
  pages={304--310},
  year={2016},
  publisher={Oxford University Press},
  url={https://pubmed.ncbi.nlm.nih.gov/26133894/}
}

@article{maschke2025The,
	author = {Maschke, Charlotte and Hadar, Peter and Zhang, Yicheng and Li, Jian and Ganjoo, Gauri and Hoopes, Andrew and Guazzo, Alessandro and Gupta, Aditya and Ghanta, Manohar and Nearing, Bruce and Silvers, Christine Tsien and Gunapati, Bharath and Thomas, Robert and Kim, Jennifer A. and Mukerji, Shibani S. and Dalca, Adrian and Zafar, Sahar and Lam, Alice D. and Mignot, Emmanuel and Westover, M Brandon},
	title = {The Brain Imaging and Neurophysiology Database: BINDing multimodal neural data into a large-scale repository},
	year = {2025},
	doi = {10.1101/2025.10.01.25337054},
	URL = {https://www.medrxiv.org/content/early/2025/10/02/2025.10.01.25337054},
	journal = {medRxiv},
}

@article{milbourn2024generation,
  title={Generation Scotland: an update on Scotland’s longitudinal family health study},
  author={Milbourn, Hannah and McCartney, Daniel and Richmond, Anne and Campbell, Archie and Flaig, Robin and Robertson, Sarah and Fawns-Ritchie, Chloe and Hayward, Caroline and Marioni, Riccardo E and McIntosh, Andrew M and others},
  journal={BMJ open},
  volume={14},
  number={6},
  pages={e084719},
  year={2024},
  publisher={British Medical Journal Publishing Group}
}


\end{document}